\documentclass[10pt,twocolumn,letterpaper]{article}

\usepackage{times}
\usepackage{epsfig}
\usepackage{graphicx}
\usepackage{amsmath}
\usepackage{amssymb}

\usepackage{diagbox}
\usepackage{amsmath}
\usepackage{multirow}
\usepackage{amssymb}
\usepackage{algorithm}
\usepackage[noend]{algpseudocode}
\usepackage{picinpar}
\usepackage{verbatim}
\usepackage{bm}
\usepackage{xurl}
\usepackage{bbm}

\DeclareMathOperator*{\argmin}{arg\,min}

\usepackage[pagebackref=true,breaklinks=true,colorlinks,bookmarks=false]{hyperref}

\newcommand{\beginsupplement}{%
        \setcounter{table}{0}
        \renewcommand{\thetable}{S\arabic{table}}%
        \setcounter{figure}{0}
        \renewcommand{\thefigure}{S\arabic{figure}}%
        \setcounter{equation}{0}
        \renewcommand{\theequation}{S\arabic{equation}}%
        \setcounter{section}{0}
        \renewcommand \thesection{S\arabic{section}}
     }

\newcommand{\LinesNumbered}{%
  \setboolean{algocf@linesnumbered}{true}%
  \renewcommand{\algocf@linesnumbered}{\everypar={\nl}}}%

\begin{document}

\title{Surrogate Model-Based Explainability Methods for Point Cloud NNs}

\author{Hanxiao Tan\\
AI Group, TU Dortmund\\
Otto-Hahn-Str. 12, 44227 Dortmund, Germany\\
{\tt\small hanxiao.tan@tu-dortmund.de}
\and
Helena Kotthaus\\
AI Group, TU Dortmund\\
Otto-Hahn-Str. 12, 44227 Dortmund, Germany\\
{\tt\small helena.kotthaus@tu-dortmund.de}
}
\date{}
\maketitle

\begin{abstract}
In the field of autonomous driving and robotics, point clouds are showing their excellent real-time performance as raw data from most of the mainstream 3D sensors. Therefore, point cloud neural networks have become a popular research direction in recent years. So far, however, there has been little discussion about the explainability of deep neural networks for point clouds. In this paper, we propose a point cloud-applicable explainability approach based on local surrogate model-based method to show which components contribute to the classification. Moreover, we propose quantitative fidelity validations for generated explanations that enhance the persuasive power of explainability and compare the plausibility of different existing point cloud-applicable explainability methods. Our new explainability approach provides a fairly accurate, more semantically coherent and widely applicable explanation for point cloud classification tasks. Our code is available at \url{https://github.com/Explain3D/LIME-3D}
\end{abstract}

\section{INTRODUCTION}
\label{intro}


Deep neural networks (DNNs) have risen on the stage of machine learning in recent years with prominent accuracy and omnipotent end-to-end learning capability. Especially in the field of computer vision, complex structured neural networks show better image recognition performance than humans. Despite their great success in industry, DNNs suffer from the trade-off between performance and explainability~\cite{Burkart2020a} due to the nonlinear model architectures. With an increasing demand for the credibility of decision-making, the studies of explainability for black-box models have received considerable critical attention. Existing research recognizes that models with high explainability play a crucial role in gaining user confidence, exposing potential biases in training data, and improving model robustness~\cite{BarredoArrieta2020}.

Several studies have found intimate connections between explainability and the safety of human life in safety-critical areas, e.g., in medicine~\cite{Esteva2019,messina2020survey,Ahsan2020} or autonomous driving~\cite{7995849,Cultrera_2020_CVPR_Workshops}. In the medical domain, decisions made by black-box models are unreliable and thus unacceptable~\cite{Samek2019}. The same dilemma occurs in the field of autonomous driving, algorithms that control vehicles with low transparency not only lead to legal problems but also pose a potential social threat~\cite{Hofmarcher2019}. Therefore, both customers and companies benefit from the research of explainable machine learning. Recent studies have proposed several explainability approaches for explaining complex machine learning models, among which the most popular methods are gradient-based~\cite{Nguyen2019,6126474,Zhou_2016_CVPR,Bach2015,shrikumar2019learning,Sundararajan2017} and local surrogate model-based~\cite{Ribeiro2016,Lundberg2017}.


On the other hand, point cloud data, as the raw data of most mainstream sensors, has a significant advantage in real-time scenarios compared to other 3D data formats and therefore has become a popular research direction in recent years. Point clouds exhibit higher structural complexity than 2D images. For instance, convolution kernels are easily applied to images due to their regularity, but they are not directly applicable to point clouds. Due to the lack of adjacency of the point cloud data, neighboring points in the point cloud matrix have a high probability of being irrelevant to the 3D spatial adjacency, which leads to the invalidation of the traditional convolution kernel. ~\cite{Qi_2017_CVPR,Qi2017+,DBLP:journals/corr/abs-1711-08488} bring up solutions for point feature extraction and make point clouds suitable to convolutional neural networks. However, in contrast to the field of 2D image processing with a large number of explainability studies~\cite{Ahsan2020,10.1007/978-3-030-33850-3_6,chen2018lshapley,vermeire2020explainable}, Most point cloud-compatible DNNs currently remain black-boxes due to the paucity of research investigating their inner explainability~\cite{9206688}. This indicates an indispensable need for the explainability research on DNNs dealing with point cloud data to ensure the transparency of decisions made by robots and autonomous vehicles, which pose a potential threat to human life.

As for the reliability examination of explainability methods, to date, there is no acknowledged evaluation criterion. Most of the previous work validates the explanation results subjectively based on human senses, which easily leads to bias in the evaluation of explainability approaches. Therefore, quantitative evaluations are increasingly recognized as an essential requirement in the explainable machine learning domain.

This work proposes a point cloud-applicable \textit{local surrogate model-based} approach~\cite{Ribeiro2016,Lundberg2017} investigating the explainability and reliability of point cloud neural networks. With the help of explanation, humans gain a better awareness of the underlying factors of misclassification cases. Besides, we quantify the plausibility of the explanations for point cloud data through fidelity and accuracy verification methods instead of a subjective approach based on human senses. Our contribution is primarily summarized as follows:

\begin{itemize}
\item We propose a local surrogate model-based explainability approach for point cloud DNNs based on LIME~\cite{Ribeiro2016}, which is more widely applicable than gradient-based methods~\cite{9206688}. 

\item We provide two quantitative evaluations for 3D explanations: fidelity metrics and cluster flipping, which are applied to validate the fidelity and plausibility of surrogate model-based and all 3D explainability approaches, respectively.

\item We evaluate our new methods and compare them against existing explainability approaches for point cloud data quantitatively using the proposed evaluation approach. Besides, we demonstrate an interesting viewpoint on the misclassified cases through our proposed approach.
\end{itemize}

The overall structure of this paper takes the form of five sections: In section \ref{related_work}, we introduce the outline of existing explainability methods and 3D neural networks, and the possibility of validating explainability approaches. Section \ref{XAI_for_PC} sets out details of our explainability approaches and the corresponding verification metrics. In section \ref{experiments}, we present the qualitative and quantitative results of our proposed methods. In section \ref{conclusion}, we conclude a brief summary and suggest future research directions.
\section{RELATED WORK}\label{related_work} 
This section reviews the current widely used explainability approaches, summarizes the classical point cloud neural network, presents existing explainability methods for point cloud DNNs, and identifies the current possibilities for verifying the explainability approaches.

\textbf{Explainability approaches:} Most of the current research on explainability pays particular attention to image classification tasks. Popular methods for explaining DNNs are gradient-based and local surrogate model-based.

\textit{Gradient-based} approaches observe the process of gradient descents during forward passes. Therefore they are only applicable to models with gradients such as neural networks. Saliency maps~\cite{simonyan2014deep} launch a pioneering attempt to explain DNNs by computing the partial derivative to each pixel of the image as its attribution. However, vanilla gradients suffers from saturated gradient~\cite{sundararajan2016gradients} and discontinuity~\cite{smilkov2017smoothgrad}. Integrated Gradient~\cite{Sundararajan2017}, Layer-wise Relevance Propagation (LRP)~\cite{Bach2015} and DeepLIFT~\cite{shrikumar2019learning} solve the saturated gradient problem by estimating the global importance of each pixel~\cite{9021958}. On the other hand, SmoothGrad~\cite{smilkov2017smoothgrad} relieves the discontinuity issue by smoothing the discontinuous gradient with a Gaussian kernel that randomly samples the input neighbors and computes their average gradients, Guided Backpropagation~\cite{springenberg2015striving} provides sharper gradient maps by removing gradients that have negative attributions to the prediction.

Another series of approaches that utilize the gradients is activation maximization~\cite{Nguyen2019}. Instead of explaining individual instances (local explanation), it attempts to discover the ideal input distribution of a given class (global explanation) by optimizing the gradients of the inputs while freezing all parameters of the networks.

\textit{Local surrogate model-based} methods such as LIME~\cite{Ribeiro2016} and KernelSHAP~\cite{Lundberg2017} aim to track the decision boundary around the selected instance by perturbing input instances and feeding them into surrogate linear models that approximate the performance of the original one but are more explainable due to the simplicity and transparency. This aims to provide a point cloud-applicable explainability approach based on local surrogate model-based methods as they are complete model-independent and applicable to arbitrary machine learning models, which offers users possibilities of more choices.

\textbf{3D convolutional neural networks}: Recent developments in the field of robotics and autonomous driving have led to an incremental interest in 3D deep learning. Processing raw point cloud data efficiently plays an important role in designing systems with low energy consumption and real-time behavior since point clouds are the main data format directly obtained from most sensors. Point clouds have higher structural complexity than 2D image data due to their disordered peculiarity, which means a lack of neighborhood consistency between data structures and spatial coordinates. The inconsistency leads to an unreproducible result when the convolution kernel is applied to raw point clouds without preprocessing. As a solution,~\cite{Maturana2015,Wu2015,Qi2016} reforms and organizes point clouds into voxels and extracts features using 3D convolution kernels.~\cite{Bruna2013,Masci2015} feed the neural networks with polygonal meshed spatial information as a substitute for the raw point clouds. However, these preprocessing approaches are not applicable in scenarios with real-time constraints and most of them are also not advantageous for semantic segmentation tasks. \cite{Qi_2017_CVPR,Qi2017+} propose point cloud-applicable convolutional networks which concatenate the local features extracted by point-wise convolutional kernels with the global feature simply obtained by max-pooling layers and achieve the state-of-the-art accuracies Modelnet40~\cite{Wu2015}, which is the currently most popular point cloud classification dataset and is also the one used in our experiments.

\textbf{Explainability in 3D DNNs}: Few studies have attempted to investigate the explainability of 3D DNNs. Although \cite{Zhang_2020} refers to explainable point cloud classification, their work addresses the disorderly properties of point clouds using PointHop Units to adapt them to classical classifiers, which is part of the pre-processing rather than post-hoc explanations. \cite{Zheng_2019_ICCV} obtains point saliency maps by simply dropping points, which is not relevant to the explainability approaches. \cite{9206688}, the pioneer study of utilizing explainability approaches to point clouds remains crucial to our understanding of feature sparsity of 3D models. However, they only show sparse explanations that emphasize the importance of points at edges and corners, which is lack-of-semantics, and the evaluation criterion of the explanations is absent. In addition, the gradient-based methods are not adapted to models without gradients, such as tree-based models. In contrast, local surrogate model-based approaches are completely model-agnostic and are applicable to arbitrary models. 

\textbf{Explanation plausibility verification:} Although there are many studies in the literature on the outcome of explainability methods, an acknowledged quantitative assessment for those approaches is absent~\cite{Burkart2020a} due to their subjectivity. \cite{adebayo2020sanity} argues that a feasible explanation should be sensitive to the weights of models and the data generating process, and proposed an alternative evaluation approach by randomizing the network weights as well as the labels and inspecting the sensitivity of the saliency maps. However, this approach tends to only benefit the gradient-based explainability methods and validates invalidity instead of feasibility. \cite{Gemert2020} strive to observe the improvement of the core performance of the network and the confidence they can generate for the users of the system when processing image data. \cite{Bach2015,7552539,Montavon2019} purpose an intuitive and efficient pattern to verify the explanations by flipping the pixels that contribute positively or negatively (or approximately to zero) to a particular class and record the verified prediction scores. Nevertheless, the flipping operation of this method could be optimized to some extent while processing point cloud data, which we will discuss in section \ref{XAI_for_PC}.
\section{EXPLAINABILITY APPROACHES FOR POINT CLOUDS}\label{XAI_for_PC}

A significant advantage of surrogate model-based methods is that they are more widely applicable. In this section, we describe in detail our explainability approach, i.e. local surrogate model-based methods for point cloud data based on LIME~\cite{Ribeiro2016}. In addition, we elaborate the quantitative evaluation approach for point cloud explanations, which evaluate the local fidelity and plausibility of existing point cloud explainability methods.

\subsection{Local surrogate model-based explainability approaches for Point Clouds} \label{explainLIMEandKS}
Local surrogate model-based explainability approaches aim to generate an explanation for a classifier $f$ and a specific instance $x$ from the data set $X$. To enable these methods for point cloud data, necessary pre-processing is obligated.

\begin{algorithm} \label{Algo1}
\caption{Pre-processing of Local surrogate model-based methods for point clouds}
\begin{algorithmic}
\Function{3D K-Means with FPS}{$P,n_c,maxIter$}
\State \textit{Input:} $P \rightarrow N \times D$ point clouds, $n_c \rightarrow$ number of clusters, $maxIter\rightarrow$Max iterations \\\#Output indicates each point belongs to which cluster
\State \textit{Output:} $C \rightarrow 1 \times N$ matrix \\\#Sample $n_c$ points from P
\State $Centers$ $\leftarrow$ FPS($c$ from $P$)
\\\#Find nearest centers for each point
\While {maxIter} :
    \For{$i$ in $n_c$}
        \State $EDMatrix$ $\leftarrow$ $\left \|P,Centers\right \|_2$
    \EndFor
    \State $minDis$ $\leftarrow$ $\argmin{(EDMatrix)}$ \\\#Point belongs to the nearest cluster, re-define the centers
    \For{$j$ in $n_c$}
        \State $P[C_j]\leftarrow$Where $minDis$ == $j$
        \State $newCenters$ $\leftarrow$ Mean($P[C_j]$)
    \EndFor
    \State $Centers$ $\leftarrow$ $newCenters$
\EndWhile
\State return $C$
\EndFunction
\end{algorithmic}
\end{algorithm}

\subsubsection{Pre-processing}
For explaining point cloud classification tasks with input size $P$ utilizing local surrogate model-based explainability approaches, each point $p \in P$ is considered as a feature individually. However, to avoid explosive computational complexity and to organize the disordered point cloud data, we group the points into super-points $C$ as features to be perturbed. We initialize a user-defined parameter $n_c$, indicating the number of the centers of clusters. To ensure uniformity and strengthen semantics we employ Farthest Point Sampling (FPS) to select $n_c$ points from $P$ and group all $p$ according to spatial coordinates using 3D K-Means Clustering such that $\forall p \in P: p \in C_i$. The pseudo-code is presented in Algorithm \ref{Algo1}.

\subsubsection{LIME for point clouds}
Same as processing 2D images~\cite{Ribeiro2016}, LIME for point cloud data also satisfies the following constraint:
\begin{equation}
    \xi(x)=argminL(f,g,\pi_x) + \Omega(g)
\end{equation}
where $f$ and $g$ denote the classifier and the explainable model for a local instance $x$ respectively, $\pi_x$ denotes the proximity measure between samples $z$ to the input $x$ (locality around $x$), and $\Omega(g)$ denotes the complexity of the explainable model. LIME tries to minimize the locality-dependent loss $argminL(f,g,\pi_x)$ by approximating $g$ to $f$. It takes samples $z$ around $x$ and feeds the perturbed samples $z'$ into $f$ to obtain a faithful surrogate model $g$ that approximates $f$, also, it regularizes the complexity of the surrogate model $g$ to guarantee that it is still explainable to humans.

Due to a large number of label categories in the point cloud dataset, explainability is hardly guaranteed even for linear surrogate classifiers. We therefore train a linear regressor that approximates the prediction score of the corresponding category from the neural network. We sample $z\in Z$ by randomly flipping component clusters from $x$ and feed the perturbed samples $Z$ into the regressor $g$ to obtain the predictions $g(z)$. To minimize $L(f,g,\pi_x)$, a kernel filters the generated samples $Z$ around $x$ based on the similarity between $z$ and $x$ proportionally (the fewer clusters being flipped the higher being weighted). The surrogate model is subsequently trained with the weighted samples $Z$ using linear regression. Due to the simplicity and transparency of linear models, it is explainable and understandable to humans and intuitive as to which parts (clusters) have positive/negative attributions to a particular prediction according to the parameters of the surrogate linear regressor.

\subsubsection{Variable input size flipping (VISF)} \label{VISFmethod}
LIME generates adjacent perturbation samples by flipping the corresponding clusters of the original instance. There are three widely-used flipping methods for a target cluster regarding 2D images: zero clearing all the included pixels, replacing those pixels with the average of the selected cluster (or the whole image), or reversing the sign of their coordinates. However, we argue that the above three operations can barely eliminate all information of the target cluster. For instance, although all pixel values are zeroed out, the contours formed by zeros remain on the data matrix and may still be learned by the neural networks. For a point cloud instance $P \in \mathbb{R}^{N \times D}$, the pixel values represent 3D spatial coordinates, and the above alternatives are likely to form a highly overlapped point set, resulting in uncertainty to determine whether the prediction fluctuations of the neural network are merely caused by flipping operations (see Fig \ref{flipcompare} for intuitive visualizations). 

To address the above problem, we simply discard the points contained in the target cluster $c_i$ from the original instance as a means to completely wipe off the information of the target cluster, i.e. $s_i = P \backslash c_i \in \mathbb{R}^{(N-\left \| c_i \right \|) \times D}$. This approach is only applicable for point cloud neural networks. Recall the architecture of point cloud networks, the final symmetric function (i.e. the max-pooling layer) addresses to extract the global feature from disordered point clouds while the local feature in the lower layer are weighted by numerous $1 \times 1$ convolutional kernels, which allows the input size of the network to be arbitrarily reformed without obstructing the inference. Notably, the variable input size flipping is both extendable in explaining and verification process (section \ref{exp_varification}).

\subsubsection{Attribution summarizing}

Local surrogate model-based methods return the individual importance of each feature, which is represented as weights of spatial coordinates in point clouds. To summarize the attribution of individual points, we simply sum up those weights:
\begin{equation}
    C_p=\sum {(C_1,C_2,C_3)}
\end{equation}

Where $C_{1\sim 3}$ stand for the attributions in each of the three spatial axes. Different summarizing patterns have varied impacts on the explanations, which is worthy of further exploration. 

\subsection{Plausibility verification for 3D explanations} \label{ver_exp}
\subsubsection{Local fidelity metrics} \label{Local fidelity metrics}
The fidelity indicates the prediction coherence between the original black-box model and the surrogate one, which is formulated as:
\begin{equation}
    F=\frac{\sum \mathbbm{1} \left (f(Z)= g(Z) \right )}{\left \| Z \right \|}
\end{equation}
Nevertheless, instead of a classifier we utilize a linear regressor as the surrogate model, which returns the prediction score only associated with the predicted class. We thus compare the batched similarity between regression scores $g(Z) \in \mathbb{R}^{\left | Z \right |}$ and the prediction scores in the corresponding logits unit of the network $f(Z)$ via several loss and coefficient measurements:
\begin{itemize}
\item Mean loss: $L_{m}=\frac{\sum_{i}^{\left \| Z \right \|}\left | f(Z)_i-g(Z)_i \right | }{\left \| Z \right \|}$
\item Mean L1 and L2 loss: $L_1=\frac{\sum\left | f(Z)-g(Z) \right | }{\left \| Z \right \|}$ 

and $L_2=\frac{\sum_{i}^{\left \| Z \right \|}(f(Z)_i-g(Z)_i)^2 }{\left \| Z \right \|}$
\item Weighted L1 and L2 loss: 

$\\L^\omega_{1}=\frac{\sum_{i}^{\left \| Z \right \|}(\left | f(Z)_i-g(Z)_i \right |\cdot \omega)}{\left \| Z \right \|}$ 

and $L^\omega_{2}=\frac{\sum_{i}^{\left \| Z \right \|}((f(Z)_i-g(Z)_i)^2\cdot  \omega)}{\left \| Z \right \|}$
\item Weighted coefficient of determination: $\\R_{\omega}^2=1-\frac{\sum_i^{\left \| Z \right \|} (f(Z)_i-g(Z)_i)^2}{\sum_i^{\left \| Z \right \|} (f(Z)_i-\overline{f_{\omega}(Z)})^2}$
\item Weighted adjusted coefficient of determination: 

$\hat{R}_{\omega}^2=1-(1-R_{\omega}^2)\left [ \frac{\left \| Z \right \|-1}{\left \| Z \right \|-\left \| g \right \|-1} \right]$ 
\end{itemize}

where $\omega$ indicates the weights derived from the kernel, $\left \| Z \right \|$ denotes the number of observed samples, $\left \| g \right \|$ is the number of parameters of $g$ and $\overline{f_{\omega}(Z)}$ indicates the weighted average. $L_{m}$,$L^{(\omega)}_{1}$ and $L^{(\omega)}_{2}$ measure the discrepancy in predicted scores while $R_{\omega}^2$ indicates the correlation between the prediction scores of the proxy model and the output of the neural network. In general, better agent approximations possess lower loss and higher decision coefficients with the predictions of neural networks. However, $R_{\omega}^2$ is sensitive to the number of samples and prone to positive bias under small sample size~\cite{doi:10.1080/02522667.1987.10698887}. We therefore introduce $\hat{R}_{\omega}^2$, which takes into account the size of variables and samples. Note that $\hat{R}_{\omega}^2$ 
has the significant range between $(-\infty, 1]$ under the assumption that $\left \| Z \right \|>\left \| g \right \|$, while the opposite case may exist in our experiments, it is therefore only informative for the case $\left \| Z \right \| > \left \| g \right \|$ in the experiment.

\subsubsection{Method-independent explanation verification} \label{exp_varification}
Fidelity metrics are only suitable for surrogate model approaches. Additional measuring methodologies are required for the reliability of non-surrogate-based explainability methods such as gradients-based saliency maps. According to the hypothesis of local accuracy of additive feature attribution ~\cite{Lundberg2017}:
\begin{equation}
   f\left ({x}  \right )=g\left ({x}'  \right )=\phi_0+\sum_{i=1}^{M}\phi_ix_i
\end{equation}
the output of original model $f\left ({x}  \right )$ is composed by linear summation of the individual feature attributions $\phi_{i}$. One of the most intuitive ways to verify an explainability approach is to eliminate features with certain attributions $\phi_{i}$ (normally positive or negative) according to their generated explanations and observe whether the output of the model $f(x)$ exhibits corresponding variations:
\begin{equation}
f\left (x  \right )-f\left (x\backslash i  \right )
\begin{cases}
 \geqslant  0 & \text{ if  }  \phi_{i}{x_i} \geqslant 0\\
 \leqslant  0 & \text{ if  }  \phi_{i}{x_i} \leqslant 0\\
\end{cases},i\in M \\
\end{equation}
where $\phi_{i}{x_i}$ denotes the attribution of flipped feature and $f\left (x\backslash i  \right )$ denotes the output of the model after flipping feature $i$. 

Nevertheless, in point cloud DNNs, the sensitivity of prediction scores for batch data is difficult to observe quantitatively due to the heterogeneous prediction scores (the logits before the softmax) from different instances. Therefore, we normalize the variability of the predicted scores to facilitate its presentation in the form of an average prediction scoreline, which is formulated as
\begin{equation}
S_{avg}=\frac{1}{n}\sum_{i=0}^{n}\frac{S_i-S_{i_{min}}}{S_{i_{max}}}
\end{equation}
where $S_i$ denotes each score in the $i$th test (including positive, negative and random perturbation series), $S_{i_{min}}$ and $S_{i_{max}}$ denote the minimum and maximum values in the corresponding evaluation run respectively.

In addition, due to the use of clustered points, we determine the averaged attributions of clusters $\phi_{c}{x_c}$ in our work rather than individual points $\phi_{i}{c_i}$, where
\begin{equation}
    \phi_{c}{x_c} = \sum_{i=1}^{c}\phi_{i}{x_i}
\end{equation}
, which lead to fluctuations in the prediction scores. The issue can be alleviated by increase the number of clusters. We discuss it further in Section \ref{Experiments}.

To quantitatively compare the plausibility among all types of explainability methods, we record the prediction scores of flipping the positive, negative and random contributing clusters respectively, denoted as $S_{pos}$, $S_{neg}$ and $S_{rdm}$. The plausibility of the corresponding explanation can be formulated as:

\begin{equation}
\bar{p}=-\frac{\sum_i^{\left \| Z \right \|} \rho_i(S_{pos}-S_{rdm},S_{neg}-S_{rdm})}{\left \| Z \right \|}    
\end{equation}

where $\rho(a,b)$ denotes the correlation coefficient between $a$ and $b$. Intuitively, flipping positive clusters results in a decline of predicted scores while flipping negative clusters lifts them up. Flipping random clusters represents the impact of eliminating neutral clusters independent of attributions, as randomly selected clusters may consist of both positive and negative points and are therefore considered indifferent. $S_{pos}-S_{rdm}$ and $S_{neg}-S_{rdm}$ are then approximations of unbiased attribution-flipping processes. A plausible explanation should have exactly opposite sensitivities to contrary attributions, and therefore its correlation coefficient of the prediction score series is expected to be as small as possible, i.e., a high score of $\bar{p}$. We consider this value as a succinct definition of the plausibility of the explainability method.
\section{EXPERIMENT}\label{experiments}
\label{Experiments}
In this section, we present the qualitative results of 3D surrogate model-based explainability methods (Sec. \ref{Qual}), evaluate and compare with other 3D-applicable approaches utilizing the quantitative verifications proposed in section \ref{ver_exp} (Sec. \ref{Quan}) and show how the explanations help to analyse the misclassified cases of the classifier (Sec. \ref{Confusing}). In our experiments\footnote{Our code is available at \url{https://github.com/Explain3D/LIME-3D}}, $1000$ test instances are selected from Modelnet40~\cite{Wu2015}, which contains $12311$ CAD models in $40$ common categories and is currently the most widely-applied point cloud classification data set. We choose PointNet~\cite{Qi_2017_CVPR} as the model to be explained, which achieves an overall accuracy of $89.2\%$ on Modelnet40. we sample $1024$ points from each instance as input to the network. Additionally, we choose Exponential Smoothing kernel for training linear regressor, denoted as
\begin{equation}
    K=\sqrt{e^{\frac{-d^{2}}{w^{2}}}}
\end{equation}
where $d$ denotes the distance from samples to the instances to be explained, and $w$ denotes the kernel width being set to a constant of $0.25$.

\begin{figure*}
\centering
\includegraphics[width=1.0\textwidth]{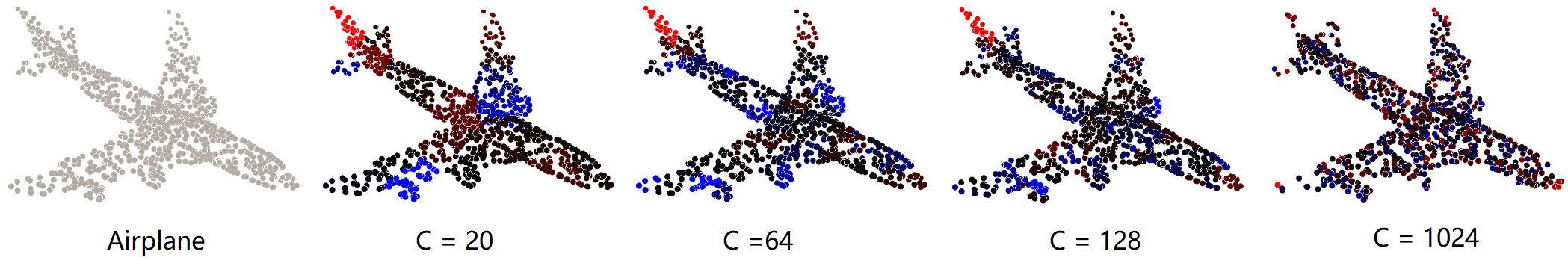}
\caption{Examples of explanations with $1000$ perturbation samples. $C$ denotes the number of clusters. Brighter red points represent more positive contributions and, conversely, brighter blue points represent more negative contributions and dim points indicate zero contributions to the corresponding classification labels.}
\label{LimeShapEx}
\end{figure*}

\subsection{Qualitative local surrogate model-based explanations} \label{Qual}
\subsubsection{Explanations visualization}
Examples of explanations generated by PointNet, as well as their original point cloud structures are shown in Figure \ref{LimeShapEx}. What stands out in the figure is that explanations with different $C$ are consistent overall except the one with $C=1024$. We believe that the reason is that $1000$ samples are insufficient for such a large number of clusters ($1024$) and therefore the surrogate model is not well-trained. Explanations based on clusters suffer from contribution neutralization. A cluster may consist of positive and negative contributing points simultaneously, aggregating them as an entity obscures the individual contribution of each point ($C=20$,$64$ and $128$). The neutralization can be alleviated by increasing the number of clusters, with the side effect of requiring more training samples and prolonged processing time ($C=1024$).

\subsection{Quantitative verification of explanation plausibility} \label{Quan}
Assessing the explanations by intuition is not quantitatively verifiable and is vulnerable to bias. This section mainly demonstrates the results of plausibility verification experiments i.e. local fidelity metrics in subsection \ref{local fidelity} and the methods-independent verification approach in subsection  \ref{point flipping}. There are two hyper-parameters for the proposed explainability method: Number of clusters $C$ and number of perturbation samples $S$. In this section, we choose $C=128$ and $S=10^3$ as the standard performance of the proposed explainability method, since $C=128$ is experimentally proved to achieve the best quantitative performance while maintaining the qualitative semantics. $S=10^3$ generates high-qualified explanations within an acceptable processing time (see table \ref{Proc.Time}) and thus is considered as the best configuration. Detailed experiments refer to hyper-parameters can be seen in Supplementary section \ref{Hyperparameters show}.

\subsubsection{Local fidelity metrics} \label{local fidelity}

Local fidelity metrics address measuring the prediction similarity between the original black-box model and the surrogate one, which play a pivotal role in verifying the plausibility of local surrogate model-based explainability methods. Due to the absence of related results as a reference, we treat the non-modified LIME (hard transplanted to point clouds) as the baseline. Table \ref{LIMEcompare} compares the local fidelity of different explaining mechanisms, i.e. whether Farthest Points Sampling (FPS) is used or whether Variable Input Size Flipping (VISF) is employed. Corresponding metric symbols refer to section \ref{Local fidelity metrics}. According to the results, our point cloud-applicable explainability approach (FPS. + VISF.) outperforms others in terms of most fidelity metrics. Note that the local fidelity measures only how closely the surrogate model approximates the black-box model, one drawback of which is that it is only applicable to explainability methods based on local surrogate models. Popular explainability methods (already proposed for point clouds) such as gradients-based ones are not compatible with these metrics, which confuses the user in choosing the most appropriate explainability method for specific tasks. 

\begin{table*}[]
\centering
\resizebox{\textwidth}{12mm}{
\begin{tabular}{cccccccc}
\hline
           & $L_{m}$               & $L_1$              & $L^\omega_{1}$       & $L_2$              & $L^\omega_{2}$       & $R_{\omega}^2$ & $\hat{R}_{\omega}^2$  \\ \hline
Base          & $1.40 \times 10^{-2}$ & $1.11 \times 10^{-1}$ & $8.66 \times 10^{-2}$ & $1.06 \times 10^{-1}$ & $6.53 \times 10^{-2}$ & 0.338          & 0.241                        \\
FPS.       & $1.22 \times 10^{-2}$ & $9.80 \times 10^{-2}$ & $7.66 \times 10^{-2}$ & $8.67 \times 10^{-2}$ & $5.35 \times 10^{-2}$ & $\bm{0.353}$          & $\bm{0.257}$                       \\
VISF.      & $1.18 \times 10^{-2}$ & $1.01 \times 10^{-1}$ & $7.90 \times 10^{-2}$ & $9.68 \times 10^{-2}$ & $5.95 \times 10^{-2}$ & 0.335          & 0.237                        \\
FPS.+VISF. & $\bm{1.03 \times 10^{-2}}$ & $\bm{8.89 \times 10^{-2}}$ & $\bm{6.95 \times 10^{-2}}$ & $\bm{7.84 \times 10^{-2}}$ & $\bm{4.82 \times 10^{-2}}$ & 0.346          & 0.249                        \\ \hline
\end{tabular}} 
\caption{Local fidelities of different explaining mechanics for point cloud data, where base denotes the base LIME explainer, FPS. denotes employing Farthest Point Sampling instead of randomly choose clusters and VISF. denotes the Variable input size flipping mechanism.}\label{LIMEcompare}
\end{table*}

\subsubsection{Methods-independent plausibility verification} \label{point flipping}
To address the aforementioned drawback we instead compare all existing point cloud-applicable explainability approaches utilizing the method-independent verification proposed in section \ref{exp_varification}. Identically, we set $C=128$ and $S=10^3$ as the "competitor" of our proposed method. Besides positive and negative attributions, We also flipping the same percentage of randomly-selected points as the baseline of prediction scores.

\begin{figure*}
    \centering
    \includegraphics[width=0.7\textwidth]{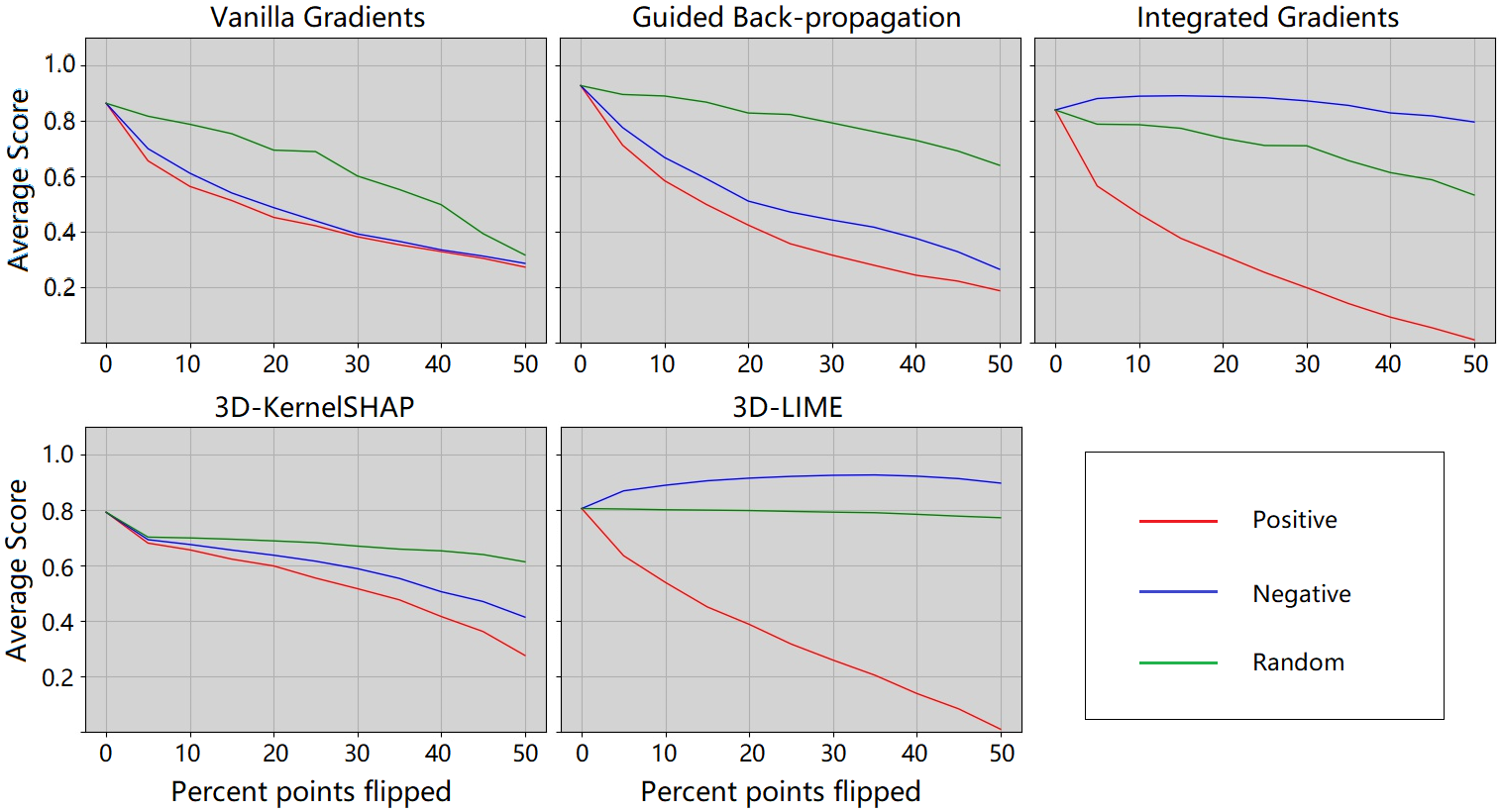}
    \caption{Variation trends of the prediction scores (y-axis) by flipping and re-inference. The scores is the average of normalized prediction scores of $1000$ test instances. Red and blue lines indicate the trend of flipping positive and negative contribution points, respectively, the green line indicates flipping random points that are independent of contribution. The x-axis indicates the percentage of flipped points for a given instance.}
    \label{QuantitativeComparison2}
\end{figure*}

\begin{table}[]
\centering
\resizebox{0.45\textwidth}{12mm}{
\begin{tabular}{cccc}
\hline
                    & $\bar{p}_{.15}$ & $\bar{p}_{.3}$ & $\bar{p}_{.5}$ \\ \hline
Vanilla Gradients       & $-0.574$        & $-0.569$       & $-0.672$       \\
Guided Back-propagation & $-0.741$        & $-0.695$       & $-0.623$       \\
Integrated Gradients    & $0.484$         & $0.366$        & $0.236$        \\
KernelSHAP              & $-0.205$        & $-0.257$       & $-0.256$            \\
LIME                    & \bm{$0.622$}         & \bm{$0.531$}        & $\bm{0.372}$        \\ \hline
\end{tabular}}
\caption{Plausibility $\bar{p}$ of flipping top-$\%15$,$\%30$ and $\%50$ attributed points.} \label{coeff table}

\end{table}

Figure \ref{QuantitativeComparison2} and Table \ref{coeff table} demonstrates the trends of prediction scores and the correlation coefficient $\bar{p}$ between different existing 3D-applicable explainability methods respectively. As the gradient-based approaches yield individual attributions of each point, we calculate coefficients for different percentages of points for fairness, i.e. top-$\%15$,$\%30$ and $\%50$ positive ones. What stands out in the results is that the explanations generated by 3D LIME and Integrated Gradients behave robustly. Their average prediction scores deteriorated rapidly after the gradual flipping of the most positive contribution points and conversely tended to increase when the negative contribution points are flipped.

On the other hand, Vanilla Gradients, Guided Back-propagation and 3D KernelSHAP are unable to distinguish between points with different contributions, resulting in gradient maps being less uniform than Integrated Gradients~\cite{9206688}. Interestingly, KernelSHAP is a variant of LIME based on Shapley value, differing from the latter solely in the choice of kernels. KernelSHAP assigns high weights to perturbation samples with individual clusters remained which severely impairs the global structure of the instance. Empirically, we find that such kernels may be more suitable for black-box model structures on other data types, but with limited performance in explaining point clouds.

We also compare the plausibility among all explaining mechanisms and the corresponding scores are presented in table \ref{plausibility tricks}. The proposed method also dominates which is consistent with the results in section \ref{local fidelity}.

\begin{figure}
    \centering
    \includegraphics[width=0.4\textwidth]{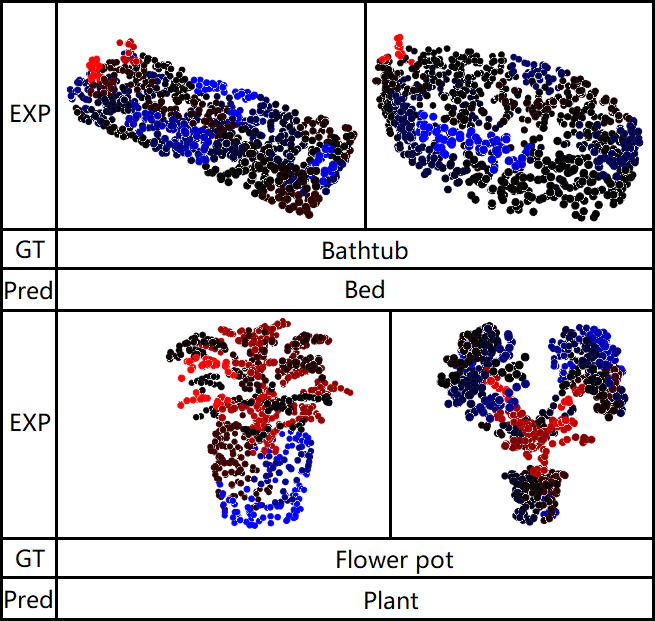}
    \caption{Explanation of the misclassification examples. Brighter red points indicate more positive contributions, while brighter blue points indicate more negative contributions and dim points indicate zero contributions. All contributions are concerning the prediction class (wrong class instead of the ground truth).}
    \label{MissclassLimeKernelS}
\end{figure}

\begin{table}[]
\centering
\begin{tabular}{cccc}
\hline
           & $\bar{p}_{.15}$ & $\bar{p}_{.3}$ & $\bar{p}_{.5}$ \\ \hline
Base       & 0.598           & 0.520          & 0.341          \\
FPS.       & 0.615           & 0.513          & 0.361          \\
VISF.      & 0.593           & 0.514          & 0.345          \\
FPS.+VISF. & $\bm{0.622}$           & $\bm{0.531}$          & $\bm{0.372}$          \\ \hline
\end{tabular}
\caption{Plausibility of different explaining mechanics for enhancing the explanation quality.} \label{plausibility tricks}
\end{table}

\subsection{Applying local surrogate model-based explainability methods for failure analysis} \label{Confusing}
A potentially applicable prospect of the local surrogate model-based explainability methods is the failure analysis. This analysis has important implications for understanding the erroneous attention paid by the classifier and has scopes for further research, e.g. 3D model revision. Figure \ref{MissclassLimeKernelS} shows examples of the attributions to the misclassified instances.

As can be seen from Figure \ref{MissclassLimeKernelS}, a majority of the misclassifications were caused by misdirected attention. In the first two examples, the faucet instead of the bathtub itself possesses the most positive attribution, misleading the model to neglect the structure of the primary target object. We believe this is because only a tiny fraction of the bathtubs in the training data is accompanied by a faucet. Another similar mislead frequently occurs with the class "Flower pot". The majority of the model's attention is drawn to the plant above rather than the pot below, resulting in a prediction of 'Plant' instead of "Flower pot" (the pot even draws a negative contribution to the ground truth label). From a human perspective, this type of data is ambiguously labeled, as both classes "Plant" and "Flower pot" are reasonable ground truth. Towards a more accurate model, this labeling type is supposed to be avoided whenever possible.

\section{CONCLUSION}\label{conclusion}
  

Although point cloud neural networks have received critical attention in recent years, so far, however, there have been few studies on their explainability. Our work proposed explainability approaches for point clouds based on local surrogate model-based methods LIME~\cite{Ribeiro2016}. We also provided the possibility to quantitatively validate the point cloud explanations. We evaluated and compared the performance of our approach against different existing explainability methods for point cloud data. The evaluation comparison revealed that our local surrogate model-based approaches as well as Integrated Gradients yield relatively plausible explanations and dominate other methods such as Guided Back-propagation. Our results also demonstrated that a larger amount of clusters and more perturbed samples are required to avoid accuracy compromise, which however consumes more processing time. Moreover, we provided intuitive analyses for misclassification cases by utilizing the proposed method. The analyses showed that part of misclassified cases is attributed to the anomalous structural distributions or ambiguous labels of the input data, misdirecting the attention of the classifier.

This work attempted to shed light on 3D neural networks. There is still tremendous potentials for further progress. Most local surrogate model based-explainability methods suffer from sample distortion as they treat each feature independently, with neither relationships nor causality present, producing unlikely combinations of features and resulting in reduced quality of explanation. Constraining the causal structure of perturbed samples to more closely resemble the training samples by introducing prior knowledge is a promising solution. On the other hand, an expected area of research is to generate more comprehensible and interesting explanations for point clouds, for instance, global explainability approaches or instance-based methods such as activation maximization or generating adversarial examples.

\clearpage
{\small
\bibliographystyle{ieee_fullname}
\bibliography{egbib}
}
\clearpage
%
\beginsupplement
\section{Supplementary materials}
These materials are supplements for the main paper. In section \ref{Hyperparameters show} we present the best hyper-parameter settings of our explainability method. In section \ref{VISF vs others}, we visually compare the proposed flip operation with the most popular ones currently available.

\subsection{Hyper-parameter configuration} \label{Hyperparameters show}
As the majority of point cloud-based classification algorithms incorporate sampling operations in their pre-processing, with $1024$ generally being the most common input size, in this section we investigate the most appropriate hyper-parameter setting for the input size $1024$. 

We sample the number of clusters with $20$, $64$, $128$ and $1024$. To ensure that $10$ sets of prediction scores are obtained while at most $50\%$ of the points are flipped in the proposed model-independent verification method, we set the number of points per cluster to approximate $5\%$ of the input size as the minimum value, i.e., $20$ clusters. As the other extreme, each point is individually regarded as a cluster, where the number of clusters is $1024$. $64$ and $128$ are randomly picked considering the processing time. We selected $10^2$,$10^{2.5}$,$10^3$,$10^{3.5}$ and $10^4$ as the incremental number of perturbation samples. Table \ref{fidelity table 1} and \ref{fidelity table 2} present the local fidelities with various hyper-parameter settings. Note that $\hat{R^2_{\omega}}$ is more convincing when comparing different $S$ because $R^2_{\omega}$ is sensitive to the number of samples $S$. When comparing different $C$, both $R^2_{\omega}$ and $\hat{R^2_{\omega}}$ can be considered as references. But since R2 is significant only for $S > C$, $R^2_{\omega}$ is chosen as a more complete observation. The results show that the proximity benefit from both greater $C$ and $S$, while there is little utility but severe time-consuming when $S$ exceeds a certain threshold ($S$ magnifies from $10^3$ to $10^4$). On the other hand, augmenting $C$ dramatically enhances the fidelity of the surrogate model due to the alleviation of attribution neutralization. However, explaining points individually suffers from loss of the semantic meaning of clusters (see figure \ref{LimeShapEx}), which is more incomprehensible to humans.

\begin{figure*}
        \centering
        \includegraphics[width=1.0\textwidth]{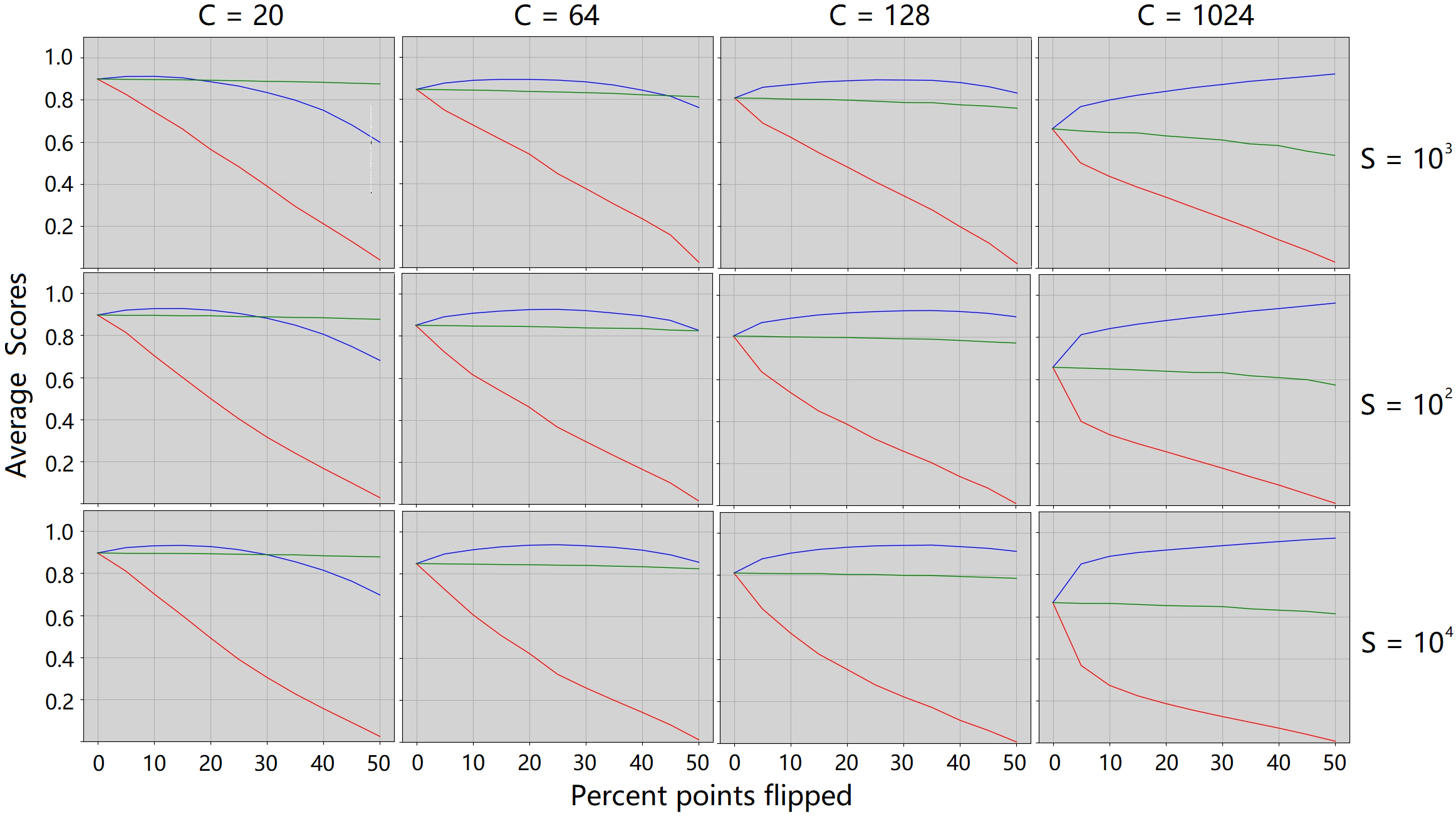}
        \caption{Means of prediction scores for different combinations of parameters of LIME. C denotes the number of clusters (features) and S denotes the number of samples used to train the surrogate model. The red and blue lines indicate the means of flipping positive and negative contributing points, while the green line indicates random flipping of the same percentage of arbitrary points.}
        \label{VariousParameters}
\end{figure*}

\begin{figure*}
        \centering
        \includegraphics[width=0.8\textwidth]{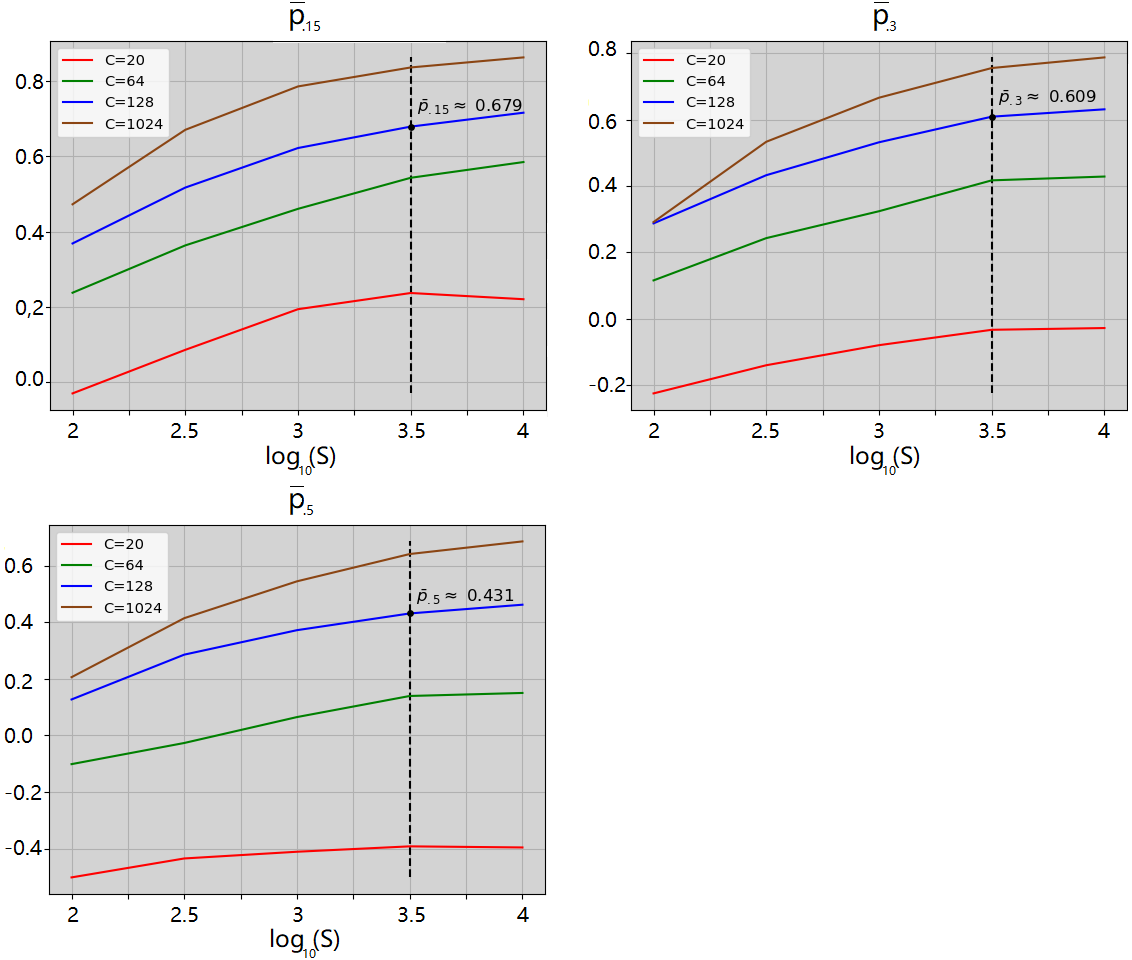}
        \caption{Plausibility with different hyper-parameter settings, i.e. $\bar{p}_{.15}$, $\bar{p}_{.3}$ and $\bar{p}_{.5}$.}
        \label{VariousParameters2}
\end{figure*}

The plausibilities under different combinations of hyper-parameters are depicted in figure \ref{VariousParameters} and \ref{VariousParameters2}. Identical to the local fidelity, the plausibility benefits from enlarging the number of clusters as well, whereas they both suffer from the semantic issue. On the other hand, enhancing the plausibility via additional perturbation samples is ineffective. At the expense of $10$ times the number of perturbed samples, only less than $0.1$ additional plausibility is achieved, which is unacceptable for point clouds with the high demand of real-time. The time costs under a various number of samples are presented in table \ref{Proc.Time}.  Since the run time of the explaining process is independent of the number of clusters, we only demonstrate its relationship to the number of perturbation samples $S$.

To summarize, we find that setting $C=128$ and $S=10^3$ is most suitable for the most popular point cloud classification model with $1024$ sampling points as input.

\begin{table}[h]
\centering
\setlength{\tabcolsep}{3mm}
\begin{tabular}{|c|c|c|c|c|c|}
\hline
S       & $10^2$ & $10^{2.5}$ & $10^3$ & $10^{3.5}$ & $10^4$ \\ \hline
Time(s) & 1.05   & 2.21       & 5.62   & 17.51      & 58.96  \\ \hline
\end{tabular}
\caption{Average processing time (in seconds) of LIME on single point cloud instance concerning the number of perturbed samples. All values are recorded as the average of $1000$ experiments.}
\label{Proc.Time}
\end{table}

\begin{table*}[]
\centering
\begin{tabular}{cccccccc}
\hline
C    & $L_{m}$               & $L_1$              & $L^\omega_{1}$       & $L_2$              & $L^\omega_{2}$       & $R_{\omega}^2$ & $\hat{R}_{\omega}^2$ \\ \hline
20   & $4.28 \times 10^{-1}$ & $7.84 \times 10^{-1}$ & $2.91 \times 10^{-1}$ & $3.02$                & $4.23 \times 10^{-1}$ & 0.239          & 0.223                \\
64   & $4.47 \times 10^{-2}$ & $1.88 \times 10^{-1}$ & $1.27 \times 10^{-1}$ & $2.77 \times 10^{-1}$ & $1.21 \times 10^{-1}$ & 0.249          & 0.198                \\
128  & $1.03 \times 10^{-2}$ & $8.90 \times 10^{-2}$ & $6.95 \times 10^{-2}$ & $7.84 \times 10^{-2}$ & $4.82 \times 10^{-2}$ & 0.345          & 0.249                \\
1024 & $\bm{1.28 \times 10^{-4}}$ & $\bm{1.00 \times 10^{-2}}$ & $\bm{8.73 \times 10^{-3}}$ & $\bm{1.73 \times 10^{-3}}$ & $\bm{1.41 \times 10^{-3}}$ & $\bm{0.883}$          & \textbackslash{}     \\ \hline
\end{tabular}
\caption{Local fidelity metrics of different $C$ with $1000$ perturbation samples.} \label{fidelity table 1}
\end{table*} 

\begin{table*}[]
\centering
\begin{tabular}{cccccccc}
\hline
S      & $L_{m}$               & $L_1$              & $L^\omega_{1}$       & $L_2$              & $L^\omega_{2}$       & $R_{\omega}^2$ & $\hat{R}_{\omega}^2$ \\ \hline
$10^2$ & $5.30 \times 10^{-2}$ & $\bm{1.86\times 10^{-1}}$  & $\bm{7.78 \times 10^{-2}}$ & $3.03 \times 10^{-1}$ & $7.59 \times 10^{-2}$ & $\bm{0.389}$          & -0.728               \\
$10^3$ & $4.47 \times 10^{-2}$ & $1.88 \times 10^{-1}$ & $1.27 \times 10^{-1}$ & $2.77 \times 10^{-1}$ & $\bm{1.21 \times 10^{-1}}$ & 0.249          & 0.198                \\
$10^4$ & $\bm{2.64 \times 10^{-2}}$ & $1.88 \times 10^{-1}$ & $1.38 \times 10^{-1}$ & $\bm{2.63 \times 10^{-1}}$ & $1.34 \times 10^{-1}$ & 0.209          & $\bm{0.204}$                \\ \hline
\end{tabular}
\caption{Local fidelity metrics of different $S$ with $64$ clusters.} \label{fidelity table 2}
\end{table*}

\subsection{VISF. VS. traditional flipping operations} \label{VISF vs others}
\begin{figure*}
        \centering
        \includegraphics[width=1.0\textwidth]{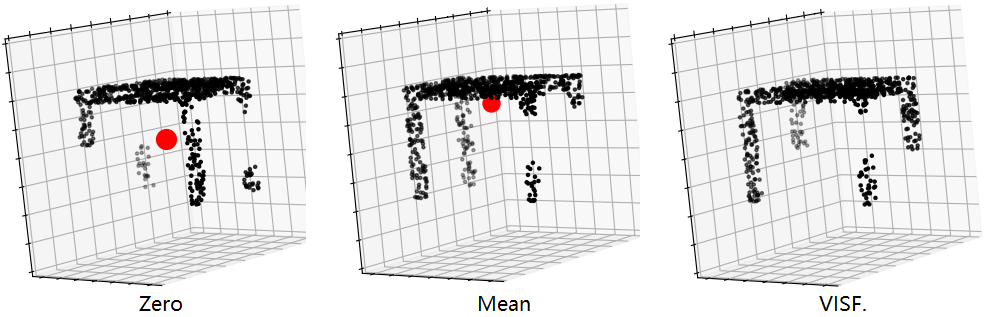}
        \caption{Visualization of three different flipping methods ($20\%$ points are flipped). Zero and mean denotes replacing the candidate points to be flipped with zeros and the mean of the three axes of rest points respectively. VISF. denotes our Variable input size flipping. The more overlapped points located at the same coordinates, the larger the diameter of that point in the image. For better observation, the points of the flipped destination are marked in red.}
        \label{flipcompare}
\end{figure*}

Fig. \ref{flipcompare} compares the three aforementioned flipping operations in section \ref{VISFmethod}. As can be seen from the figure, the currently most popular flipping operations (i.e. replacing the candidate points to be flipped with zeros and means of the rest points) fail to eliminate additional interference of flipping. As there is a lumpy collection of overlapping points at a specific location, it is difficult to determine whether the variations in predicted scores are independent of this lump of points. In contrast, the instances flipped by VISF. contain no similar lump, leading to a more convincing association of predicted scores with important features.

\end{document}